\documentclass{acm_proc_article-sp}

\usepackage{graphicx}
\usepackage{subfigure} 
\usepackage{amsmath}
\usepackage{rotating}
\usepackage{multirow}
\usepackage{multicol}
\usepackage[toc,page]{appendix}
\usepackage{longtable}
\usepackage{algorithm}
\usepackage{algorithmic}
\usepackage[table]{xcolor}
\usepackage{footnote}

\newcommand{\myem}[1]{\textcolor{blue}{\bf #1}}

\begin{document} 

%

\title{Combining One-Class Classifiers Using Meta Learning}


\numberofauthors{3}
\author{
\alignauthor Eitan Menahem\\
		\affaddr{Deutsche Telekom Laboratories}\\
		\affaddr{Department of Information Engineering}\\
		\affaddr{Ben-Gurion University of the Negev}\\
		\affaddr{Be'er Sheva, 84105, Israel}\\
		\email{eitanme@post.bgu.ac.il}\\
\alignauthor Lior Rokach\\
		\affaddr{Deutsche Telekom Laboratories}\\
		\affaddr{Department of Information Engineering}\\
		\affaddr{Ben-Gurion University of the Negev}\\
		\affaddr{Be'er Sheva, 84105, Israel}\\
		\email{liorr@post.bgu.ac.il}\\
\alignauthor Yuval Elovici\\
		\affaddr{Deutsche Telekom Laboratories}\\
		\affaddr{Department of Information Engineering}\\
		\affaddr{Ben-Gurion University of the Negev}\\
		\affaddr{Be'er Sheva, 84105, Israel}\\
		\email{elovici@post.bgu.ac.il}\\
}

\maketitle
\begin{abstract}
Selecting the best classifier among the available ones is a difficult task, especially when only instances of one class exist. In this work we examine the notion of combining one-class classifiers as an alternative for selecting the best classifier. In particular, we propose two new one-class classification performance measures to weigh classifiers and show that a simple ensemble that implements these measures can outperform the most popular one-class ensembles. Furthermore, we propose a new one-class ensemble scheme, TUPSO, which uses meta-learning to combine one-class classifiers. Our experiments demonstrate the superiority of TUPSO over all other tested ensembles and show that the TUPSO performance is statistically indistinguishable from that of the hypothetical best classifier.
\end{abstract}

\section{Introduction and Background}

In regular classification tasks we aim to classify an unknown instance into one class from a predefined set of classes. One-class classification aims to differentiate between instances of class of interest and all other instances. The one-class classification task is of particular importance to information retrieval tasks \cite{manevitz2002one}. Consider, for example, trying to identify documents of “interest” to a user, where the only information available is the previous documents that this user has read (i.e. positive examples), yet another example is citation recommendation, in which the system helps authors in selecting the most relevant papers to cite, from a potentially overwhelming number of references \cite{bethard2010should}. Again, one can obtain representative positive examples by simply going over the references, however it would be hard to identify typical negative examples (the fact that a certain paper is not cited by another paper does not necessarily indicate it is irrelevant).
Many one-class classification algorithms have been investigated \cite{Bishop94noveltydetection,
scholkopf2001estimating,
DBLP:conf/simbad/KontorovichHM11}. While there are plenty of learning algorithms to choose from, identifying the one that performs best in relation to the problem at hand is difficult. This is because evaluating a one-class classifier's performance is problematic. By definition, the data collections only contain one-class examples and thus, performance metrics, such as false-positive ($FP$), and true negative ($TN$), cannot be computed. In the absence of $FP$ and $TN$, derived performance metrics, such as classification accuracy, precision, among others, cannot be computed. Moreover, prior knowledge concerning the classification performance on some previous tasks may not be very useful for a new classification task because classifiers can excel in one dataset and  fail in another, i.e., there is no consistent winning algorithm.

This difficulty can be addressed in two ways. The first option is to select the classifier assumed to perform best according to some heuristic estimate based on the available positive examples (i.e., $TP$ and $FN$). The second option is to train an ensemble from the available classifiers. To the best of our knowledge, no previous work on selecting the best classifier in the one-class domain has been published and the only available one-class ensemble technique for diverse learning algorithms is the fixed-rule ensemble, which in many cases, as we later show, makes more classification errors when compared to a random-selected classifier.
	
In this paper we search for a new method for combining one-class classifiers. We begin by presenting two heuristic methods to evaluate the classification performance of one-class classifiers. We then introduce a simple heuristic ensemble that uses these heuristic methods to select a single base-classifier. Later, we present TUPSO, a general meta-learning based ensemble, roughly based on the Stacking technique \cite{Wolpert92} and incorporates the two classification performance evaluators. We then experiment with the discussed ensemble techniques on forty different datasets. The experiments show that TUPSO is by far the best option to use when multiple one-class classifiers exist. Furthermore, we show that TUPSO's classification performance is strongly correlated with that of the actual best ensemble-member. 
	
\subsection{One-Class Ensemble}
The main motivation behind the ensemble methodology is to weigh several individual classifiers and combine them to obtain a classifier that outperforms them all. Indeed, previous work in supervised ensemble learning shows that combining classification models produce a better classifier in terms of prediction accuracy \cite{DBLP:journals/isci/MenahemRE09}.
	
Compared to supervised ensemble learning, progress in the one-class ensemble research field is limited \cite{GiacintoPRR08}.  Specifically, the Fix-rule technique was the only method which was considered for combining one-class classifiers \cite{Tax01combiningone-class, JuszczakD04,SeguiIV10}. In this method, the combiner regards each participating classifier's output as a single vote upon which it applies an aggregation function (a combining rule), to produce a final classification. In the following few years, further research was carried out and presently there are several applications reaching domains, such as information security (intrusion detection), remote sensing, image retrieval, image segmentation, on-line signature verification, and fingerprint matching. 
Table \ref{tab:references} summarizes the relevant work.
	
\begin{table}[t]
\centering

\begin{tabular}{|l@{\hspace{15mm}}|l|}
\hline
\textbf{Domain} & \multicolumn{1}{c|}{\textbf{Works}} \\
\hline
Theory and methods	& \cite{Tax01combiningone-class}, \cite{JuszczakD04}, \cite{ShiehK09} and \cite{SeguiIV10} \\
Information security & \cite{PerdisciGL06}, \cite{CabreraGM08} and \cite{GiacintoPRR08} \\
Remote sensing	& \cite{Munoz-MariCGC07} \\
Image retrieval	& \cite{WuC09c} \\
Image segmentation	& \cite{Cyganek10} \\
Signature verification	& \cite{Nanni06d} \\
Fingerprint matching	& \cite{NanniL06g} \\
\hline
\end{tabular}
\caption{\label{tab:references} Research in one-class ensemble}
\end{table}

In the context of the one-class ensemble the \emph{majority voting} technique is used by Perdisci et al. \cite{PerdisciGL06} and by Wu and Chung \cite{WuC09c}. Perdisci et al. shows that a classifier ensemble makes their network-based intrusion detection system (NIDS) more secure against adversary attacks, as it is more difficult for the attacker to evade all of the base-classifiers (e.g., by mimicry attacks), at the same time. Wu and Chung use the ensemble of one-class classifiers for image retrieval; they split the images into multiple instances, on which a set of weak classifiers (1-SVM) are trained separately by using different sub-features extracted from the instances. They conclude that one-class ensembles can boost the image retrieval’s accuracy and improve its generalization performance.

In addition to \emph{majority voting}, other combining rules, such as \emph{max-rule}, \emph{average-rule} and \emph{product-rule}, are seen in the literature. These, unlike majority voting, output a continuous numeric value, denoted as $y(x)$, which can be viewed as an analogue to the classification confidence score. Five similar rules are proposed by Tax \cite{Tax01combiningone-class} :  \emph{mean-votes}, \emph{mean weighted vote}, \emph{product of the weighted votes}, \emph{mean probabilities} and \emph{product combination of the estimated probabilities}. These rules can be divided into two groups: aggregates on discrete votes and aggregates on confidence scores (sometimes implemented as posterior probabilities). The first group uses the indicator function over thresholded estimated probabilities, whereas the computation in the second group of rules is performed directly on the estimated probabilities. 

The \emph{mean-votes} rule computes the average positive votes. This rule is used by Giacinto et al. \cite{GiacintoPRR08} for service-specific intrusion detection in computer networks. The proposed IDS learns multiple one-class traffic models (one model per network service group), to model the normal traffic of specific network service groups, such as web services, mail services, etc. Later, in order to combine the multiple traffic models, they use the mean-votes rule. 

The \emph{max rule} was used for on-line signature verification \cite{Nanni06d}, in which the author proposed to induce $k$ different one-class classifiers on
$k$ modified train sets and then combine the classifiers via the max-rule; each modified train set contains a random subset of the original features, as suggested in \cite{ho1998random}. The proposed method was shown to outperform state-of-the-art work for both random and skilled signature forgeries. The same combining rule was also used for fingerprint matching by Nany and Lumini in \cite{NanniL06g}. The authors apply the Random Bands technique on the fingerprint images dataset to derive $k$ new train-sets, upon which $k$ one-class classifiers are trained and then combined.

The \emph{exclusive-voting} rule was applied for cluster images \cite{Cyganek10}. This rule determines that the ensemble classifies positive only if a single member of the ensemble classifies positive, as part of their technique for image segmentation (a technique in compute vision). The author clusters the input images into $k$ groups upon which they later train $k$ one-class classifiers, referred to as expert 1-SVMs (one-class SVM). In the classification phase, the expert 1-SVMs classify an input image as 'positive' or as 'negative'. Finally, the ensemble outputs positive only if exactly one expert OC-SVMs votes positive. 

While the fix-rule ensemble techniques are very popular they have a tendency of generating non-optimal ensembles, because the combining rules are assigned \emph{statically} and \emph{independently} of the training data. As a consequence, as we will show later, the fix-rule ensembles produce an inferior classification performance in comparison to the best classifier in the ensemble.

In the following lines we use the notation $P_k(x|\omega_{Tc})$ as the estimated probability of instance $x$ given the target class $\omega_{Tc}$, $fr_{(T,k)}$ as the fraction of the target class, which should be accepted for classifier $k=1 \dots R$, $N$ as number of features, and $\theta_k$  notates the classification threshold for classifier $k$. A list of fixed combining rules is presented in Table \ref{Tab:FixCombiningRules}.

\begin{table}[h]
  
  \centering
	\resizebox{1.0\linewidth}{!} {
  \begin{tabular}{@{\hspace{2mm}}p{0.22\columnwidth}@{\hspace{1mm}}|@{\hspace{1mm}}p{1.2\columnwidth}@{\hspace{2mm}}}
    \noalign{\hrule height 1pt}
		 
     \multicolumn{1}{@{\hspace{2mm}}p{55pt}@{\hspace{1mm}}|@{\hspace{1mm}}}{\textbf{Combining Rule}} & \multicolumn{1}{c@{\hspace{2mm}}}{\multirow{2}{*}{\textbf{Combination Rule Formula}}} \\
   	\noalign{\hrule height 1pt}
    {Majority \newline voting}	& \multirow{2}{*}{$y(x)=I_{\geq{k/2}}\left(\sum_{k}I\left(P_{k}\left(x|\omega_{Tc}\right)\geq\theta_{k}\right)\right)$}\\
    \hline
    {Mean \newline vote} &	\multirow{2}{*}{$y(x)=\frac{1}{R}\sum_{k=1}^R I \left(P_{k}\left(x|\omega_{Tc}\right)\geq\theta_{k}\right)$} \\
		\hline
    {Weighted \newline mean vote}	& \multirow{2}{*}{$y(x)=${$\frac{1}{R}\sum_{k=1}^R\left[f_{T,k}I\left(P_{k}\left(x|\omega_{Tc}\right)\geq\theta_{k}\right)+(1-f_{T,k})I\left(P_{k}\left(x|		 \omega_{Tc}\right)\geq\theta_{k}\right)\right]$}}\\
    \hline
    {Average \newline rule}	& \multirow{2}{*}{$y(x)=\frac{1}{R}\sum_{k=1}^R P_{k}\left(x|\omega_{Tc}\right)$} \\
    \hline
    {Max \newline rule}& \multirow{2}{*}{$y(x)=argmax_{k}[P_{k}\left(x|\omega_{Tc}\right)]$} \\
    \hline
    {Product \newline rule}& \multirow{2}{*}{$y(x)=\prod_{k=1}^R[P_{k}\left(x|\omega_{Tc}\right)]$} \\
    \hline
		
    {Exclusive \newline voting} & \multirow{2}{*}{$y(x)=I_{1}\left(\sum_{k}I\left(P_{k}\left(x|\omega_{Tc}\right)\geq\theta_{k}\right)\right)$} \\ 
		\hline
		Weighted \newline vote product& \multirow{2}{*}{$y(x)=\frac{\prod_{k=1}^R[fr_{(T,k)}I\left(P_{k}\left(x|\omega_{Tc}\right)\geq\theta_{k}\right)]}{\prod_{k=1}^R[f_{T,k}I\left(P_{k}\left(x|\omega_{Tc}\right)\geq\theta_{k}\right)]+\prod_{k=1}^R[(1-f_{T,k})I\left(P_{k}\left(x|\omega_{Tc}\right)<\theta_{k}\right)]}$}\\[3ex]
		\noalign{\hrule height 1pt}
  \end{tabular}
	}
	\caption{Fix combining rules}\label{Tab:FixCombiningRules}
\end{table}

Instead of using the fix-rule (e.g., weighting methods), technique to combine one-class classifiers, the meta-learning approach can be used. 
\subsection {Meta Learning}	
\label{sec:metaLearning}
Meta-learning is the process of learning from basic classifiers (ensemble members); the inputs of the meta-learner are the outputs of the ensemble-member classifiers. The goal of meta-learning ensembles is to train a meta-model (meta-classifier), which will combine the ensemble members' predictions into a single prediction. To create such an ensemble, both the ensemble members and the meta-classifier need to be trained. Since the meta-classifier training requires already trained ensemble members, these must be trained first. The ensemble members are then used to produce outputs (classifications), from which the meta-level dataset (meta-dataset) is created.
 The basic building blocks of meta-learning are the meta-features, which are measured properties of the ensemble members output, e.g., the ensemble members' predictions. A vector of meta-features and a classification $k$ comprise a meta-instance, i.e., meta-instance $\equiv <f^{meta}_{1},\dots,f^{meta}_{k},y>$, where $y$ is the real classification of the meta-instance that is identical to the class of the instance used to produce the ensemble members' predictions. A collection of meta-instances comprises the meta-dataset upon which the meta-classifier is trained. 

\subsection{Estimating The Classification Quality}
Traditional classifier evaluation metrics, such as true negative and false positive, cannot be computed in the one-class setup, since only positive examples exist. Consequently, measures, such as a classifier's accuracy, precision, AUC, F-score, and Matthew’s correlation coefficient (MCC), cannot be computed, since $accuracy=(TP+\textbf{TN})/(TP+\textbf{TN}+\textbf{FP}+FN)$, $Precision=TP/(TP+\textbf{FP})$ and $F$-$score=2*\textbf{P}*R/(\textbf{P}+R)$, where $P$ is precision and $R$ is recall. Instead of computing the aforementioned metrics, Liu et al. \cite{DBLP:conf/icml/LiuLYL02} and Lee and Liu \cite{DBLP:conf/icml/LeeL03}, proposed heuristic methods for estimating, rather than actually measuring, the classifier's accuracy and F-score, respectively. 
Next, we describe the two performance estimators. 

\subsection*{One-Class Accuracy}
Liu et al. demonstrated in \cite{DBLP:conf/icml/LiuLYL02} that by rewriting the error probability, one can estimate the classification error-rate, in the one-class paradigm, given a prior on the target-class: 
\[
\resizebox{\linewidth}{!} {
$Pr[f(x)\neq y]=Pr[f(x)=1]-Pr[Y=1]+2Pr[f(x)=0|Y=1]Pr[Y=1]$
}
\] 
where $f(x)$ is the classifier's classification result for the examined example $x$, $Pr[f(x)=1]$ is the probability that the examined classifier will classify \emph{Positive}, $Pr[f(x)=0|Y=1]$ is the probability that the classifier will classify \emph{Negative} when given a \emph{Positive} example, and lastly, $Pr[Y=1]$ is the prior on the target-class probability.

Naturally, we define the one-class accuracy (OCA), estimator as follows: $OCA = 1-Pr[f(x)\neq y]$. Note that the probabilities $Pr[f(x)=1]$ and $[f(x)=0|Y=1]$ can be estimated for any one-class problem at hand using a standard cross-validating procedure.  

\subsection*{One-Class F-Measure}
An additional performance criteria, $\frac{r^2}{Pr[f(x)=1]}$, denoted as One-Class F-score (OCF), is given by Lee and Liu in \cite{DBLP:conf/icml/LeeL03}. Using this criteria, one can estimate the classifier's F-score in the semi-supervise paradigm. However, when only positive-labeled instances exist, the recall, $r=Pr[f(x)=1|y=1]$, equals to $Pr[f(x)=1]$ (because $Pr[y=1]=1$), which only measures the fraction of correct classifications on positive test examples, i.e., true-positive rate (TPR). Using the TPR to measure the classification performance makes sense, because the TPR is strongly correlated with the classification accuracy when negative examples are very rare, such as in the case of most one-class problems.

The contributions of this work are four fold: (1) our work introduces a new meta-learning-based one-class ensemble, TUPSO, which generally outperforms both fix-rule ensembles and performs on par with the best ensemble-member, (2) it presents a comprehensive comparison between multiple one-class ensembles (some are presented for the first time in the context of one-class), on 40 cross-domains data collections, (3) we adapt two unsupervised learning algorithms, namely Peer Group Analysis \cite{Eskin02ageometric} and Global Density Estimation \cite{DBLP:conf/kdd/KnorrN97}, into one-class algorithms, and (4) we investigate two metrics presented by Lee and Liu \cite{DBLP:conf/icml/LeeL03} for estimating the one-class classifier’s performance and show their usefulness for choosing the best available classifier, and for improving the one-class ensembles accuracy, with both meta-learning and weighting methods.

The rest of the paper is organized as follows. In Section ‎\ref{sec:newEnsembles} we present two novel one-class ensembles schemes: \emph{Best classifier By Estimation}, a simple heuristic based ensemble, and \emph{TUPSO}, a Stacking-like meta-learning based one-class ensemble scheme. In Section ‎\ref{sec:methods} we describe the methods and conditions in which the experiments were conducted. Later, in Section ‎\ref{sec:results} the experimental result are presented and discussed. Lastly, our conclusions and future work are presented in Section ‎\ref{sec:conclusions}.

\section{New Ensembles Schemes}
In this section we introduce two novel one-class ensemble schemes: Best-Classifier by Estimation and TUPSO.

\label{sec:newEnsembles}
\subsection{Best-Classifier By Estimation} 

Using the discussed classification performance estimators, we define a new and very simple ensemble: Estimated Best-Classifier Ensemble (ESBE). This ensemble is comprised of an arbitrary number of one-class ensemble-members (classifiers). During the prediction phase, the ensemble's output is determined by a single ensemble-member, denoted as the \emph{dominant classifier}. The ensemble's dominant member is selected during the training phase. This is achieved by evaluating the performance of the participating ensemble-members using a 5x2 cross-validation procedure, as described in \cite{DBLP:journals/neco/Dietterich98}. During this procedure only the training-set's instances are used, and the metric used to measure the ensemble-members` performance is either OCA or OCF. 

\subsection{TUPSO}
\label{sec:TUPSO}
In this section, we introduce the TUPSO ensemble scheme. The main principle of TUPSO is combining multiple and possibly diverse one-class classifiers using the meta-learning technique. TUPSO is roughly based on the Stacking technique, and as so, it uses a single meta-classifier to combine the ensembles` members. As opposed to Stacking, however, where the meta-classifier trains directly from the ensemble-members` outputs, TUPSO's meta-classifier trains on a series of aggregations from the ensemble-members` outputs. To elevate the effectiveness of some of the aggregations used by TUPSO, and with that improve the ensemble's overall performance, during the training phase, the ensemble-members are evaluated using the aforementioned one-class performance evaluators. The performance estimates are then translated into static weights, which the meta-learning algorithm uses during the training of the meta-classifier, and during the prediction phases.

\begin{figure}
	\centering
		\includegraphics[scale=0.45]{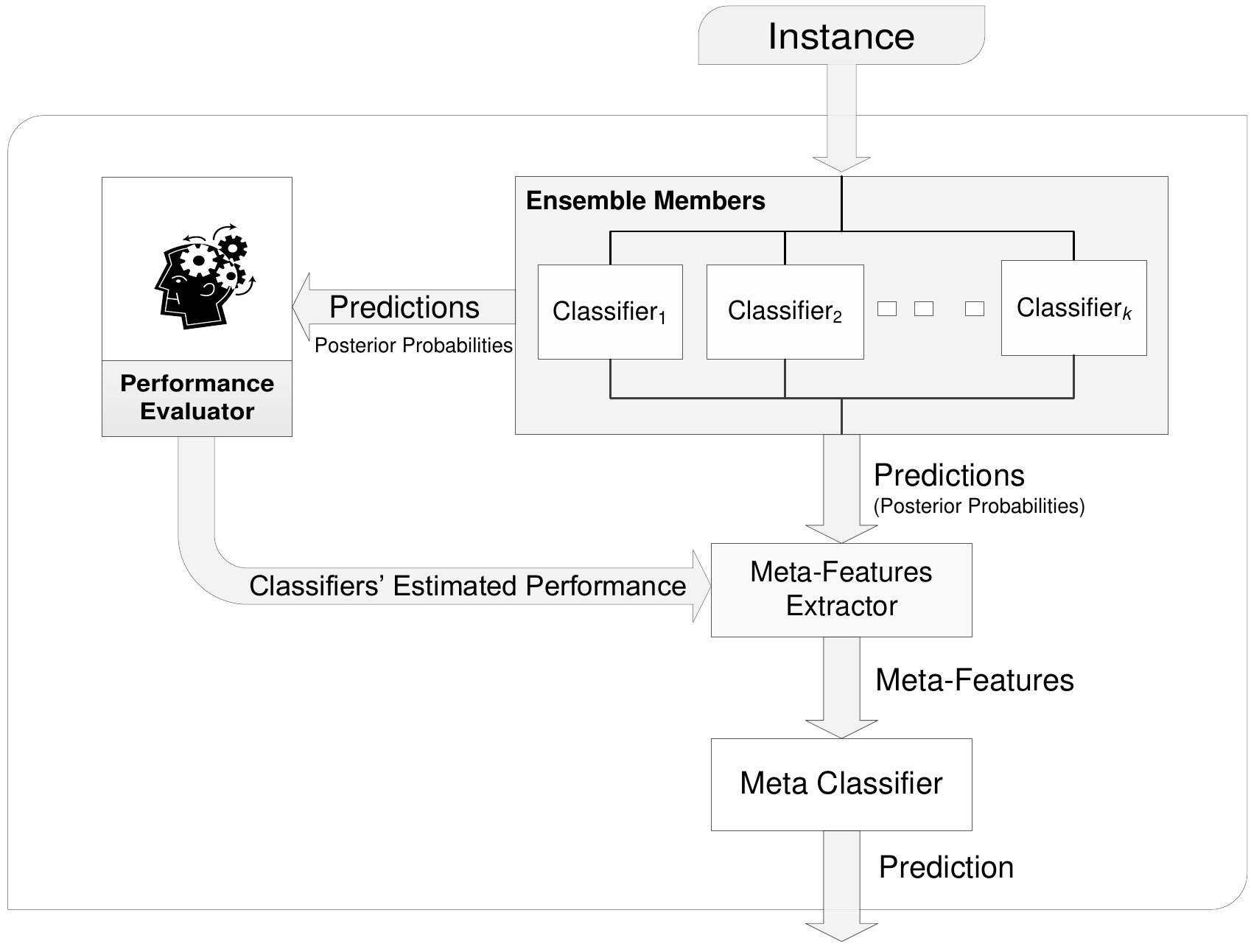}
	\caption{The TUPSO ensemble scheme.}	\label{fig:TUPSO Architecture BW}
\end{figure}

The TUPSO ensemble, as shown in Figure \ref{fig:TUPSO Architecture BW}, is made up of four major components: (1) Ensemble-members, (2) Performance evaluator, (3) Meta-features extractor, and (4) Meta-classifier. Next, we describe each component.

\subsubsection*{Ensemble Members}
In TUPSO, the ensemble members are one-class, machine-learning-based, classifiers. TUPSO regards its ensemble members as black boxes, in order to avoid any assumption regarding their inducing algorithm, data structures or methods for handling missing values and categorical features. During the ensemble's training phase, the ensemble-members are trained several times, as part of a cross-validation process, which is required for generating the meta-classifier's dataset. This process is described later in Section \ref{sec:methods}.

\subsubsection*{Performance Evaluator}
The Performance Evaluator estimates the ensemble members' classification performance during the ensemble's training phase. To fulfill its task, the Performance Evaluator uses one of the available classification performance estimators, i.e., OCA or OCF.

\subsubsection*{Meta-Features Extractor}
The meta-features are measured properties of one or more ensemble-members' output. A collection of meta features for a single instance makes a meta-instance. A collection of meta-instances is called a meta-dataset. The meta-dataset is used to train the meta-classifier. The Meta Features Extractor computes the meta-features by using multiple aggregations of the ensemble-members` output. Let  $P_m=<p_{(m_1 )},\dots,p_{(m_k)}>$  be the vector containing the ensemble-members` outputs $p_{(m_1)},\dots,p_{(m_k)}$, where $k$ is the number of members in the ensemble. A set of aggregate features is computed for each instance in the training set. A single set makes a single meta-instance, which will later be used either as a training instance for the meta-learner or as a test meta-instance. 

\begin{table}[t]
\resizebox{\linewidth}{!} {
  \centering
  \footnotesize 
  \begin{tabular}{@{}l@{\hspace{1mm}}|@{\hspace{1mm}}l@{}}
    \noalign{\hrule height 1pt}
    \multirow{2}{*}{\textbf{Meta-Feature Name}} & \multicolumn{1}{@{\hspace{3mm}}c@{}}{\multirow{2}{*}{\textbf{Aggregation Function}}} \\
     & \\
		\noalign{\hrule height .2pt}
		\multirow{2}{*}{Sum-Votes} 							&  \multirow{2}{*}{$f_{1}\left(P_{m}\right)=\sum_{i=1}^{k}1_{\left\{p_{m_{i}}\geq0.5\right\}}\left(P_{m_{i}}\right)$}\\
    & \\
		\hline
		\multirow{2}{*}{Sum-Predictions}  			&  \multirow{2}{*}{$f_{2}\left(P_{m}\right)=\sum_{i=1}^{k}P_{m_{i}}$}\\
    & \\
		\hline
		\multirow{2}{*}{Sum-Weighted-Predictions} &  \multirow{2}{*}{$f_{3}\left(P_{m}\right)=\sum_{i=1}^{k}\alpha_{i}*P_{m_{i}}$}\\
    & \\
		\hline
		\multirow{2}{*}{Sum-Power-Weighted-Predictions}	&  \multirow{2}{*}{$f_{4}\left(P_{m}\right)=\sum_{i=1}^{k}\alpha_{i}*(P_{m_{i}})^2$}\\
		& \\
		\hline
		\multirow{2}{*}{Sum-Log-weighted-Predictions} 		&  \multirow{2}{*}{$f_{5}\left(P_{m}\right)=\sum_{i=1}^{k}\alpha_{i}*\log(P_{m_{i}})$}\\
    & \\
		\hline
		\multirow{2}{*}{Var-Votes} 											&  \multirow{2}{*}{$f_{6}\left(P_{m}\right)=Var(1_{\left\{p_{m_{i}}\geq0.5\right\}}\left(P_{m_{i}}\right))$}\\
    & \\
		\hline
		\multirow{2}{*}{Var-Predictions}							  &  \multirow{2}{*}{$f_{7}\left(P_{m}\right)=Var\left(P_{m}\right)$}\\
    & \\
		\hline
		\multirow{2}{*}{Var-Weighted-Predictions} 				&  \multirow{2}{*}{$f_{8}\left(P_{m}\right)=Var\left(\alpha*P_{m}\right)$}\\
    & \\
		
    \noalign{\hrule height 1pt}
  \end{tabular}
	}
		\caption{Aggregate functions which generate TUPSO's meta features. }
	\label{tab:AggregateFeatures}
\end{table}

Table \ref{tab:AggregateFeatures} defines eight experimental aggregate meta-features. The aggregate functions  $f_2 \dots f_5$ and $f_6 \dots f_8$ are based on the first and second moments, respectively. The first moment computes the ``average'' ensemble-members` prediction, whereas the second moment computes the variability among the ensemble-members` predictions. 
The first moment based aggregation, a subtle version of the mean voting rule, is motivated by \emph{Condorcet's Jury Theorem}, and is used in several supervised-learning ensembles, e.g., Distribution-Summation \cite{clark1991rule}. Furthermore, the second moment based aggregation is motivated by the knowledge it elicits over the first moment, i.e., the level of consent among the ensemble-members. From this information, unique high-level patterns of ensemble members' predictions can be learned by the meta-learner, and thereafter be at the disposal of the meta-classifier.
Table \ref{tab:meta-dataset} shows the resulted structure of TUPSO's meta-dataset. 
\begin{table}[h]
\footnotesize
	\resizebox{\linewidth}{!} {
	\centering
	
		\begin{tabular}{@{}c@{\hspace{.5mm}}|c@{\hspace{1mm}}|@{\hspace{1mm}}c@{\hspace{1mm}}|@{\hspace{0.5mm}}c@{\hspace{0.5mm}}|@{\hspace{1mm}}c@{\hspace{1mm}}|@{\hspace{1mm}}c|c@{}}
		\hline
		Instance & $f_{1}\left(P_{m}\right)$ & $f_{2}\left(P_{m}\right)$ & $f_{3}\left(P_{m}\right)$ & ... & $f_{7}\left(P_{m}\right)$ & $f_{8}\left(P_{m}\right)$ \\
		\hline
		1 & $ma_{1,1}$ & $ma_{1,2}$ & $ma_{1,3}$ & ... & $ma_{1,7}$ & $ma_{1,8}$\\
		2 & $ma_{2,1}$ & $ma_{2,2}$ & $ma_{2,3}$ & ... & $ma_{2,7}$ & $ma_{2,8}$\\
		... & ...  & ... & ... & ... & ... & ...\\
		\hline
		\end{tabular}
		}
	\caption{The training-set of the meta-classifier. Each column represents an aggregate feature over the ensemble members' predictions, and $ma_{i,j}$ denotes the value of meta-feature $j$ for meta-instance $i$}
	\label{tab:meta-dataset}
	\vspace{-3mm}
\end{table}


\subsubsection*{Meta-Classifier}
The meta-classifier is the ensemble's combiner, thus, it is responsible for producing the ensemble's prediction. Similar to the ensemble-members, the meta-classifier is a one-class classifier; it learns a classification model from meta-instances, which consist of meta-features. Practically, the meta-features used in training the meta-classifier can be either aggregate features, raw ensemble-members` predictions or their combination. 

However, experiments, which were unable to be included in this paper\footnote{see: http://www.filedropper.com/tupso}, showed that training the meta-classifier using the raw ensemble-members` predictions alone or alongside the aggregate meta-features yielded less accurate ensembles.

\subsubsection{Training Process}
The training process of TUPSO begins with training the ensemble-members, followed by training the meta-classifier. The ensemble-members and the meta-classifier are trained using an inner $k$-fold cross-validation training process. First, the training-set is partitioned into $k$ splits. Then, in each fold, the ensemble-members are trained on $k$-1 splits. Afterwards, the trained ensemble-members classify the remaining split to produce meta-instances. The meta-instances in each fold are added to a meta-dataset. After $k$ iterations, the meta-dataset will contain the same amount of instances as the original dataset. Lastly, the ensemble-members are re-trained using the entire training-set and the meta-classifier is trained using the meta-dataset.

\subsubsection{Weighting the Ensemble Members}
In order to calculate certain meta-features, e.g., $f_3$, the ensemble-members` predictions have to be weighed. To do so, a set of weights, one per ensemble-member, are learned as part of the ensemble training process. 
During the meta-classifier training, the ensemble-members predict the class of the evaluated instances. 
The predictions are fed to the Performance Evaluator, which calculates either OCA or OCF estimations for each of the ensemble-members, $Perf_{vect}=<Perf_1,\dots ,Perf_m>$, where $Perf_i$ is the estimated performance of ensemble-member$_i$. 
Finally, a set of weights, $\alpha_1,\alpha_2,\cdots,\alpha_m$, is computed as follows:
\[\alpha_i=\frac{Perf_i}{\Sigma_{j=1}^m Perf_j}, \forall i=1 \dots m\]

\section{Methods}
\label{sec:methods}

We now specify the methods and conditions in which we investigated the presented ensemble schemes. First, we indicate the ensemble-members that participate in the ensembles. Next, we discuss the evaluated ensemble schemes. 

\subsection{One-Class Learning Algorithms}
For evaluation purposes, we made use of four, one-class algorithms: OC-GDE, OC-PGA, OC-SVM, \cite{Schlkopf99estimatingthe}, and ADIFA \cite{ADIFA}. We selected these ensemble-members because they represent the prominent families of one-class classifiers, i.e., nearest-neighbor (OC-GDE, OC-PGA), density (ADIFA), and boundary (OC-SVM). The first two algorithms are our adaptations of two well-known supervised algorithms to one class learning.

\textbf{Peer Group Analysis} (PGA), is an unsupervised anomaly detection method proposed by Eskin et al. \cite{Eskin02ageometric}, that identifies the low density regions using the nearest neighbors. An anomaly score is computed at point $x$ as a function of the distances from $x$ to its $k$ nearest neighbors. Although the PGA is actually a ranking technique applied to clustering problems, we implemented it as a one-class classifier with $k_{nn}=1$. Given the training sample $S$, a test point $x$ is classified as follows. For each $x_i\in S$, we pre-compute the distance to $x_i$'s nearest neighbor in $S$, given by $d_i=d(x_i, S \setminus \{x_i\})$. To classify $x$, the distance to the nearest neighbor of $x$ in $S$, $d_x=d(x,S)$, is computed. The test point $x$ is classified as an anomaly if $d_x=d(x,S)$ appears in a percentile $p_\alpha$ or higher among the $\{d_i\}$; otherwise, it is classified as normal. We set the parameter $p_\alpha$ to $0.01$. 

\textbf{Global Density Estimation} (GDE), proposed by \cite{DBLP:conf/kdd/KnorrN97}, is also an unsupervised density-estimation technique using nearest neighbors. Given a training sample $S$ and a real value $r$, one computes the anomaly score of a test point $x$ by comparing the number of training points falling within the $r$-ball $B_r(x)$ about $x$ to the average of $|B_r(x_i)\cap S|$ over every $x_i\in S$. We set $r$ to be twice the sample average of $d(x_i,S\setminus \{x_i\})$ to ensure that the average number of neighbors is at least one. To convert GDE into a classifier, we need a heuristic for thresholding anomaly scores. We chose the following one, as it seemed to achieve a low classification error on the data: $x$ is classified as normal if $\exp(-((N_r(x)-\bar N_r)/\sigma_r)>1/2$, where $N_r$ is the number of $r$-neighbors of $x$ in $S$, $\bar N_r$ is the average number of $r$-neighbors over the training points and $\sigma_r$ is the sample standard deviation of the number of $r$-neighbors.

We used a static pool of six ensemble-members for all the evaluated ensembles: (i) ADIFA$_{HM}$, (ii) ADIFA$_{GM}$, (iii) OC-GDE, (iv) OC-PGA, OC-SVM$_1$, and (vi) OC-SVM$_2$. The ensemble-members properties, illustrated in Table \ref{tab:ensemble-membersetup}, were left unchanged during the entire evaluation. 
\begin{table}[h!]
	\resizebox{1\linewidth}{!} {
	\begin{tabular}{ll@{\hspace{2mm}}l}

    \noalign{\hrule height 1pt}
    \small Base Classifier & \small Algorithm & \small Parameters\\
    \noalign{\hrule height 1pt}
    ADIFA$_{HM}$ &	$ADIFA$ & $\Psi=Harmonic Mean, s=2\%$ \\
    ADIFA$_{GM}$ &	$ADIFA$ & $\Psi=Geometric Mean, s=1\%$\\
    OC-GDE &	$OC$-$GDE$ & $n/a$\\
    OC-PGA &	$OC$-$PGA$ & $k=3$, $p_\alpha=0.01$\\
    OC-SVM$_1$&	$OC$-$SVM$ & $k=linear$, $\nu=0.05$\\
    OC-SVM$_2$&	$OC$-$SVM$ & $k=polynomial$, $\nu=0.05$\\
    \noalign{\hrule height 1pt}

  \end{tabular}}	
	\caption{ensemble-members setup parameters. The non-default parameters are illustrated.}
	\label{tab:ensemble-membersetup}
\end{table}

\subsection{Ensemble Combining Methods}
\label{sec:methods}
The following evaluation includes several ensemble combining methods from three groups of algorithms: Heuristic-Ensemble: estimated best-classifier ensemble (ESBE); Fixed-rules: \emph{majority voting, mean-voting, max-rule} and \emph{product-rule}; and Meta-learning-based: \emph{TUPSO}. The learning algorithm used for inducing the meta-classifier in TUPSO was ADIFA, as it outperformed the other three mentioned learning algorithms on the evaluation set.

\subsection{Datasets} 
During the evaluation we used a total of 40 distinct datasets from two different collections, UCI and KDD-CUP99. All datasets are fully labeled and binary-class. 

We selected 34 datasets from the widely used UCI dataset repository \cite{blake1998uci}. The datasets vary across dimensions, number of target classes, instances, input features, and feature type (nominal or numeric). So as to have only two classes in the UCI datasets, a pre-process was completed where only the instances of the two most prominent classes were selected. The other instances were filtered out.

The \emph{KDD CUP 1999} dataset contains a set of instances that represent connections to a military computer network. The dataset contains 41 attributes, 34 of which are numerical and 7 of which are categorical. The original dataset contained 4,898,431 multi-class data instances. In order to divide the dataset into multiple binary-class sets, we followed the method performed in \cite{DBLP:conf/kdd/YamanishiTWM00}. Compared with the UCI datasets, the KDD99-CUP are much more \emph{natural} one-class datasets, as they are highly imbalanced (instances of the network's normal state make the lion's share of the derived binary datasets). 

\subsection{Measured Metrics}
We used the area under the ROC curve (AUC) metric to measure the classification performance of the individual classifiers and ensemble methods. The ROC (Receiver Operating Characteristic) curve is a graph produced by plotting the TPR versus the fraction of false positives (FPR) for a binary classifier as its discrimination threshold varies. The AUC value of the best possible classifier will be equal to 1, indicating that we can find a discrimination threshold under which the classifier will obtain 0\% false positives and 100\% true positives. 

\subsection{Evaluation Methodology}
During the training phase, only the examples of one-class were available to the learning algorithms and to the classification performance estimators. During the testing phase, however, both positive and negative examples were available, to evaluate the classifiers in real-life conditions. The generalized classification accuracy was measured by performing a 5x2 cross-validation procedure \cite{DBLP:journals/neco/Dietterich98}. In each of the cross-validation iterations, the dataset was randomly partitioned into two disjoint instance subsets. In the first fold, the first subset was utilized as the training set, while the second subset was utilized as the testing set. In the second fold, the role of the two subsets was switched. This process was repeated five times. The same cross-validation folds were implemented for all algorithms.

To conclude which ensemble performs best over multiple datasets, we followed the procedure proposed by Dem{\v{s}}ar in \cite{DBLP:journals/jmlr/Demsar06}. In the case of multiple ensembles of classifiers or features, we first used the adjusted Friedman test so as to reject the null hypothesis, followed by the Bonferroni-Dunn test to examine whether a specific ensemble or feature produces significantly better AUC results than the reference method.

\section{Experimental Results}
\label{sec:results}
In this section we examine the performance of the discussed ensembles. Our ultimate goal is to learn which ensemble, if any, can make the task of selecting the best classifier redundant. 
Since some of the examined ensembles are based on heuristics, i.e., ESBE and TROIKA, we first measure the correlation between these heuristics and their parallel two-class metrics, i.e. OCA vs. Accuracy, and OCF vs. F-Measure. If the heuristic and their parallel two-class metrics are strongly correlated, then the heuristics performance measurements can be effectively used as a substitute for the unavailable two-class performance measurements.
In the second experiment, we examine our one-class classifiers, to ensure that not a single classifier classifies significantly better than the rest of the classifiers. In other words, we want to show that selecting the best one-class classifiers in not a trivial task.
Next, in the third experiment we compare the classification performance of the existing (i.e. fix-rule), one-class ensembles with the new one-class ensembles proposed in this work.
Lastly, we determine whether an ensemble of classifiers performs as good as the actual best ensemble-member classifier. 

Note the difference between the \emph{actual}, best ensemble member and the \emph{estimated}, best ensemble member. The first is determined during the evaluation phase, where both positive and negative instances exist, whereas the second is computed during the training phase, where only positive instances exist, and therefore, we should expect it to be inferior to the \emph{actual}, best ensemble-member.

In the statistical analysis tables further ahead, we use the following notations; the `+' (`-') symbol indicates that the average AUC value of the ensemble, indicated at the row's beginning, is significantly higher (lower), compared to the ensemble indicated in the table’s header with a confidence level of 95\%. 

\subsection{One-Class and Two-Class Performance \newline Metrics Are Correlated}

\begin{figure*}[t]
	\begin{minipage}[l]{0.5\linewidth}
	\centering
		\includegraphics* [scale=0.4]{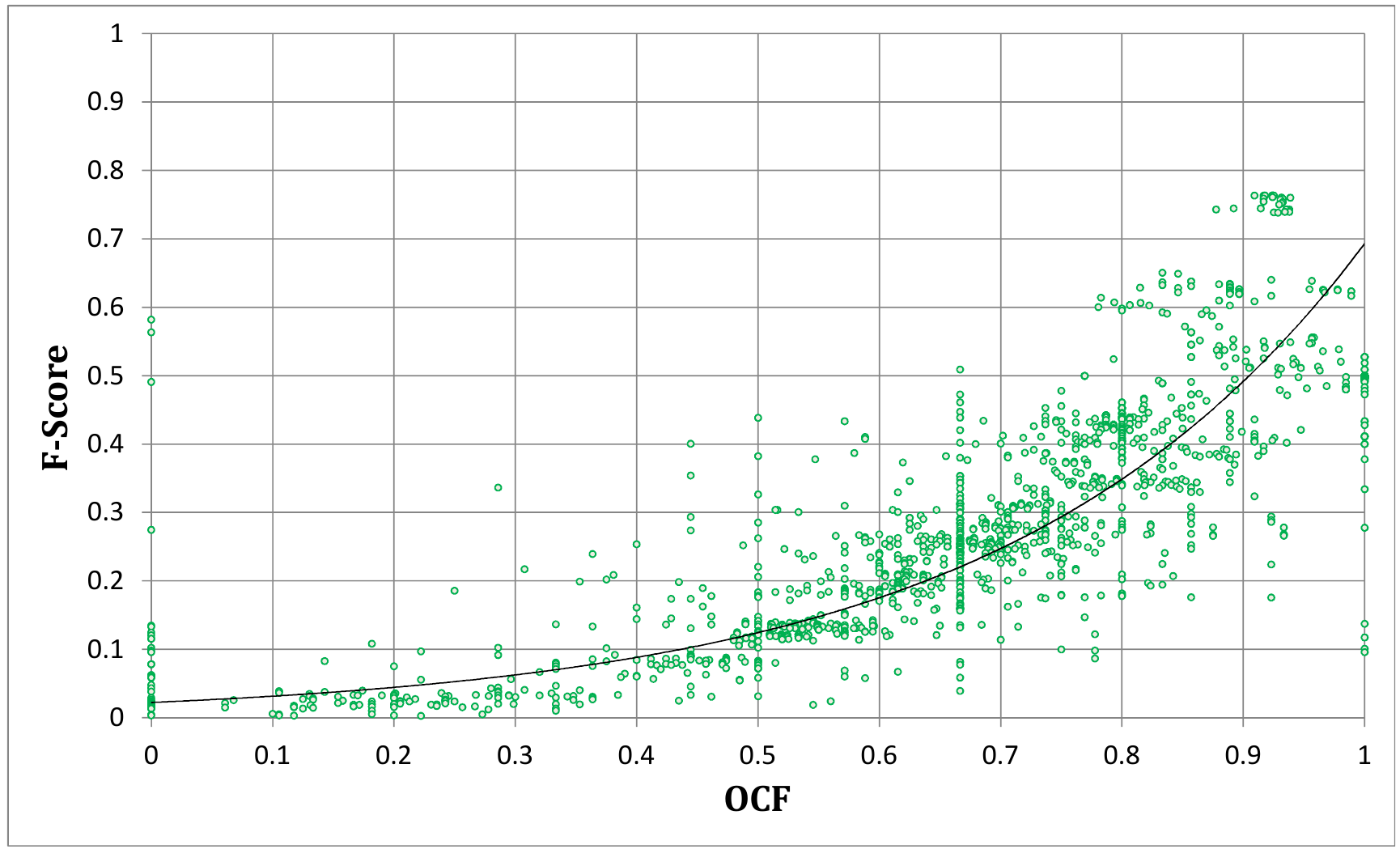}
	\caption{F-Measure vs. OCF}
	\label{fig:Correlation2}

\end{minipage}
\begin{minipage}[r]{0.5\linewidth}
	\centering
		\includegraphics* [scale=0.4]{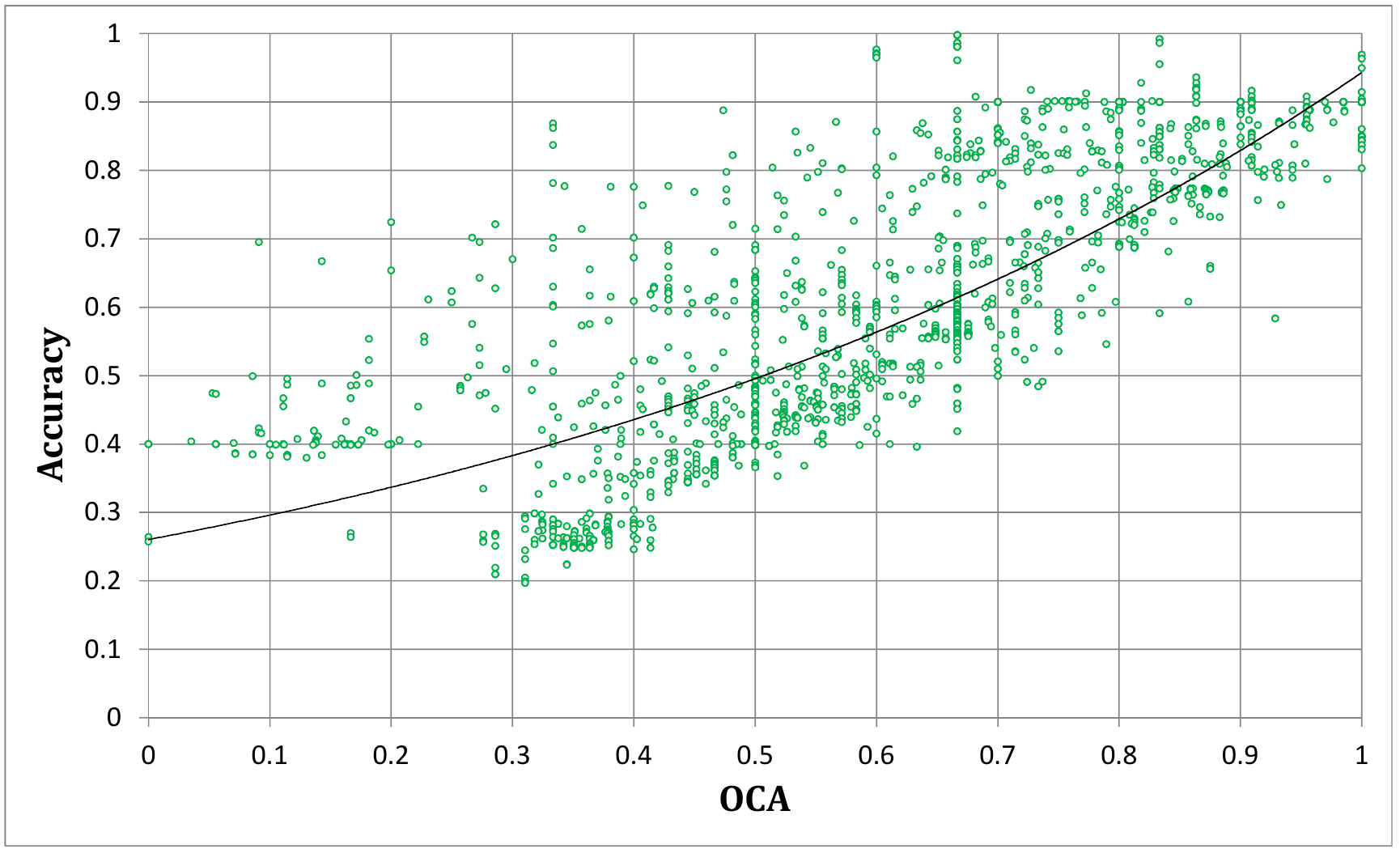}
	\caption{Accuracy vs. OCA}
	\label{fig:Correlation3}
\end{minipage}
\end{figure*}

To investigate the relationship between the two associated performance evaluators in question, i.e., Accuracy vs. OCA and F-Score vs. OCF, we trained six one-class classifiers, as described in Table \ref{tab:ensemble-membersetup}, on the UCI collection datasets. Next, the classifiers were evaluated and the relevant performance metrics (OCA, OCF, TPR, F-measure, and Accuracy), were measured. To measure the first three measures, only the test-set’s positive examples were used, whereas to compute the last two metrics, both positive and negative examples were used. With these performance measures a correlation matrix was calculated, presented in Table \ref{tab:CorrelationBetweenOneAndTwoClassPerformanceMetrics}. In addition, Figures \ref{fig:Correlation2} and \ref{fig:Correlation3} show two correlation plots; the first plots the F-measure versus the OCF graph, whereas the second plots the Accuracy versus the OCA graph.

\begin{table}[h]
	\centering
		\resizebox{1.0\linewidth}{!} {
		\begin{tabular}{lc@{\hspace{9mm}} c @{\hspace{9mm}} c@{\hspace{9mm}} c @{\hspace{9mm}} c}
		\noalign{\hrule height 1pt}
									& F-measure & OCF 	& Accuracy 	& OCA 		& TPR \\
																																			\noalign{\hrule height 0.5pt}
			F-measure 	& 1 					& - 					& - 					& - 			& -	  \\
			OCF 				& \myem{0.87} & 1 					& - 					& - 			& - 	\\
			Accuracy 		& 0.92	 			& 0.78				& 1	 					& - 			& - 	\\
			OCA 				& 0.61	 			& 0.58				& \myem{0.81}	& 1 			& - 	\\
			TPR 				& 0.38	 			& 0.24 				& 0.16	 			&  -0.27	& 1 	\\
	\noalign{\hrule height 1pt}
		\end{tabular}
		}
	\caption{Correlation Between one and two-class performance metrics}
	\label{tab:CorrelationBetweenOneAndTwoClassPerformanceMetrics}
\end{table}

Table \ref{tab:CorrelationBetweenOneAndTwoClassPerformanceMetrics} shows a sturdy positive correlation between OCA and Accuracy (0.81) and even a slightly greater correlation between OCF and F-measure (0.87). Interestingly, OCF and Accuracy are also highly correlated, which might be attributed to the inherent high correlation between F-Measure and Accuracy measures.

\subsection{No Clear Winner Base-Classifier}
The goal of the next experiment is to verify that picking up the best-classifier is not a trivial task. In other words, we need to ensure that there is no constant or very probable winner classifier. Otherwise, an ensemble of classifiers cannot be justified, since picking the best classifier could be simply done via a cross-validation procedure. 

We calculate the \emph{entropy}, an uncertainty measure, of the classifiers average ranking over a series of dataset to examine how predictable the best-classifier is. When there is a very probable winner classifier, the calculated entropy is very small (in particular, zero for a constant winner classifier). In the opposing case, where all classifiers have the same average rank, the calculated entropy will be at its maximal peak.

The participating classifiers were executed on the \emph{UCI} and \emph{KDD-CAP} datasets. The classification results, and their classifiers relative ranks are presented in Table \ref{tab:BaseClassifiersResultTable}. Table \ref{tab:ClassifiersRankStatistics} summarizes the classifiers ranking and presents the per-classifier entropy and the per-rank entropy statistics. 	

\begin{table}[b!]
	\centering
		\resizebox{1\linewidth}{!} {
		\begin{tabular}{@{}|l|p{0.31\linewidth}@{}p{0.21\linewidth}@{}p{0.21\linewidth}@{}p{0.19\linewidth}@{}p{0.19\linewidth}@{}p{0.19\linewidth}@{}p{0.19\linewidth}@{}|}
\noalign{\hrule height 1pt}
														& &	 ADIFA$_{GM}$	&	ADIFA$_{HM}$	&	GDE &	PGA &	OC-SVM$_1$ &	OC-SVM$_2$ \\
\noalign{\hrule height 1pt}
\multirow{34}{*}{\begin{sideways}UCI Repository\end{sideways}}
	& Anneal 					& 0.543 (4) & 0.600 (3) & 0.865 (2) & 0.905 \myem{\myem{(1)}} & 0.489 (5) & 0.486 (6) \\ 
	& Audiology 			& 0.860 (3) & 0.881 (2) & 0.887 \myem{(1)} & 0.659 (5) & 0.611 (6) & 0.691 (4) \\ 
	& Balance-scale 	& 0.819 (2) & 0.733 (3) & 0.989 \myem{(1)} & 0.683 (4) & 0.523 (5) & 0.523 (6) \\ 
	& Breast-cancer 	& 0.491 (5) & 0.503 (3) & 0.409 (6) & 0.500 (4) & 0.528 (2) & 0.543 \myem{(1)} \\ 
	& C-Heart-disease & 0.554 (3) & 0.487 (6) & 0.704 \myem{(1)} & 0.640 (2) & 0.488 (4) & 0.488 (4) \\ 
	& Credit-rating 	& 0.689 (3) & 0.712 (2) & 0.790 \myem{(1)} & 0.639 (4) & 0.501 (5) & 0.497 (6) \\ 
	& E-coli 					& 0.928 (4) & 0.931 (3) & 0.920 (5) & 0.850 (6) & 0.936 \myem{(1)} & 0.936 \myem{(1)} \\ 
	& Heart-Statlog 	& 0.546 (3) & 0.497 (6) & 0.733 \myem{(1)} & 0.643 (2) & 0.497 (4) & 0.497 (4) \\ 
	& Hepatitis 			& 0.624 (4) & 0.686 (2) & 0.784 \myem{(1)} & 0.639 (3) & 0.476 (6) & 0.480 (5) \\ 
	& Horse-Colic 		& 0.561 (2) & 0.608 \myem{(1)} & 0.553 (3) & 0.544 (4) & 0.516 (6) & 0.522 (5) \\ 
	& Heart-Disease   & 0.650 (3) & 0.540 (4) & 0.813 \myem{(1)} & 0.704 (2) & 0.505 (5) & 0.504 (6) \\ 
	& Hypothyroid 		& 0.493 (2) & 0.632 \myem{(1)} & 0.467 (6) & 0.476 (5) & 0.485 (4) & 0.489 (3) \\ 
	& Ionosphere 			& 0.789 (6) & 0.796 (5) & 0.909 \myem{(1)} & 0.906 (2) & 0.847 (3) & 0.830 (4) \\ 
	& Iris 						& 0.935 (5) & 0.915 (6) & 0.970 \myem{(1)} & 0.970 \myem{(1)} & 0.950 (3) & 0.950 (3) \\ 
	& Chess 					& 0.52 (4) & 0.669 (2) & 0.867 \myem{(1)} & 0.641 (3) & 0.501 (6) & 0.503 (5) \\ 
	& Letter 					& 0.825 (4) & 0.880 (3) & 0.954 \myem{(1)} & 0.949 (2) & 0.482 (6) & 0.482 (5) \\ 
	& MFeature 				& 0.986 \myem{(1)} & 0.943 (2) & 0.707 (5) & 0.630 (6) & 0.775 (4) & 0.851 (3) \\ 
	& Mushroom 				& 0.606 (4) & 0.722 (3) & 1.000 \myem{(1)} & 0.744 (2) & 0.486 (5) & 0.484 (6) \\ 
	& Opt-Digits 			& 0.877 (4) & 0.941 (2) & 0.993 \myem{(1)} & 0.936 (3) & 0.648 (6) & 0.655 (5) \\ 
	& Page-Blocks 		& 0.759 (4) & 0.861 (2) & 0.889 \myem{(1)} & 0.843 (3) & 0.628 (5) & 0.627 (6) \\ 
	& Pen Digits 			& 0.984 \myem{(1)} & 0.954 (3) & 0.977 (2) & 0.954 (4) & 0.875 (5) & 0.874 (6) \\ 
	& Diabetes 				& 0.497 (3) & 0.493 (4) & 0.439 (6) & 0.488 (5) & 0.619 (2) & 0.627 \myem{(1)} \\ 
	& P-Tumor 				& 0.995 \myem{(1)} & 0.983 (2) & 0.936 (5) & 0.922 (6) & 0.975 (3) & 0.972 (4) \\ 
	& Segment 				& 0.581 (4) & 0.702 \myem{(1)} & 0.697 (2) & 0.661 (3) & 0.527 (5) & 0.525 (6) \\ 
	& Sonar 					& 0.459 (5) & 0.458 (6) & 0.537 (4) & 0.550 (3) & 0.556 (2) & 0.572 \myem{(1)} \\ 
	& Soybean 				& 0.544 (3) & 0.530 (4) & 0.859 \myem{(1)} & 0.680 (2) & 0.465 (5) & 0.457 (6) \\ 
	& SPAM-Base 			& 0.616 (3) & 0.500 (4) & 0.743 \myem{(1)} & 0.691 (2) & 0.477 (5) & 0.476 (6) \\ 
	& Splice 					& 0.985 (2) & 0.954 (3) & 0.989 \myem{(1)} & 0.553 (4) & 0.474 (5) & 0.474 (5) \\ 
	& Vehicle 				& 0.538 (4) & 0.614 (3) & 0.811 \myem{(1)} & 0.731 (2) & 0.477 (5) & 0.477 (5) \\ 
	& Vote 						& 0.896 (2) & 0.924 \myem{(1)} & 0.790 (3) & 0.451 (6) & 0.699 (5) & 0.715 (4) \\ 
	& Vowel 					& 0.606 (4) & 0.658 (3) & 0.923 (2) & 0.931 \myem{(1)} & 0.456 (5) & 0.456 (5) \\ 
	& Waveform 				& 0.748 (4) & 0.755 (3) & 0.661 (6) & 0.747 (5) & 0.852 \myem{(1)} & 0.850 (2) \\ 
	& W.B. Cancer & 0.976 \myem{(1)} & 0.919 (2) & 0.904 (3) & 0.886 (4) & 0.467 (5) & 0.455 (6) \\ 
	& Zoo 						& 0.910 (3) & 0.928 (2) & 0.988 \myem{(1)} & 0.556 (4) & 0.519 (5) & 0.494 (6) \\ 
	\noalign{\hrule height 1pt}
	\multirow{6}{*}{\begin{sideways}KDD CAP 99\end{sideways}}
	& AUTH & 0.985 \myem{(1)} & 0.950 (2) & 0.919 (4) & 0.928 (3) & 0.849 (5) & 0.849 (5) \\ 
	& FTP & 0.552 (4) & 0.720 (3) & 0.932 \myem{(1)} & 0.925 (2) & 0.487 (5) & 0.487 (5) \\ 
	& FTP-DATA & 0.663 (3) & 0.861 \myem{(1)} & 0.773 (2) & 0.654 (4) & 0.489 (6) & 0.489 (5) \\ 
	& OTHER & 0.991 \myem{(1)} & 0.955 (2) & 0.930 (3) & 0.928 (4) & 0.575 (5) & 0.575 (5) \\ 
	& POP3 & 0.993 \myem{(1)} & 0.951 (2) & 0.925 (4) & 0.934 (3) & 0.721 (5) & 0.721 (5) \\ 
	& SMTP & 0.740 (4) & 0.898 (2) & 0.863 (3) & 0.922 \myem{(1)} & 0.729 (5) & 0.729 (5) \\ 
\noalign{\hrule height 1pt}
\multicolumn{2}{|l}{Average Rank} & \multicolumn{1}{c}{3.1 (3)} & \multicolumn{1}{c}{2.85 (2)} & \multicolumn{1}{c}{\myem{2.4 (1)}} & \multicolumn{1}{c}{3.3 (4)} & \multicolumn{1}{c}{4.5 (5)} & \multicolumn{1}{c|}{4.53 (6)} \\
\noalign{\hrule height 1pt}

		\end{tabular}
		}
	\caption{Ensemble-members performance results. Inside the parenthesis is the classifiers' AUC rank.}
	\label{tab:BaseClassifiersResultTable}
\end{table}

The results in Tables \ref{tab:BaseClassifiersResultTable} and \ref{tab:ClassifiersRankStatistics}, present an interesting picture. It seems that some classifiers perform better than others, but there is no constant winner.
The entropy-per-classifier statistic, of all six classifiers, is very high, and in fact is only slightly smaller than the  maximal possible entropy value \footnote{the maximal possible entropy is obtained when all the outcomes of a random variable have the same probability, i.e., in our case $entropy_{max}=-6*(\frac{1}{6})\dot Log_2 (\frac{1}{6})=2.585$}. These results imply that the rank of a classifier is very unpredictable, hence, given the six classifiers; it will be very difficult to correctly guess their ranking, given a new dataset. Do we have, at least, a good chance of successfully predicting the best classifier? The very high per-rank-entropy (2.161), reflects the high uncertainty in the identity of the best classifier, and therefore, the answer is \emph{no}.


\begin{table}[h!]
	\centering
		\resizebox{1\linewidth}{!} {
		\begin{tabular}{p{60pt}cccccc | c}
		\hline
									& \multicolumn{6}{c|}{Rank} & Classifier \\ \cline{2-7}
									& $1^{st}$	& $2^{nd}$ 	& $3^{rd}$ 	& $4^{th}$ 	& $5^{th}$ 	& $6^{th}$ & entropy \\
		\hline
		ADIFA$_{GM}$	&	7 	& 5 	& 10 	& 14 	& 3 	& 1  & 2.258 \\
		ADIFA$_{HM}$	&	5 	& 14 	& 12 	& 4 	& 1 	& 4  & 2.224 \\
		GDE 					&	20 	& 5 	& 5 	& 3 	& 3 	& 4  & 2.143 \\
		PGA 					&	4 	& 10 	& 8 	& 10 	& 4 	& 4  & 2.461 \\
		OC-SVM$_1$ 		&	2 	& 3 	& 3 	& 4 	& 21 	& 7  & 2.037 \\
		OC-SVM$_2$	 	&	4 	& 1 	& 3 	& 6 	& 14 	& 12 & 2.207 \\
		\hline
		Rank entropy & 2.161	& 2.235	& 2.398	& 2.363	& 1.979	& 2.292 & \\
		\hline
		\end{tabular}
		}
	\caption{Classifiers rank statistics}
	\label{tab:ClassifiersRankStatistics}
\end{table}

\subsection{One-Class Ensembles Performance}
In the following experiment we examine the classification performance of the above-mentioned one-class ensembles. 
The ensembles` classification performance results are presented in Table \ref{tab:EnsembleResultTable}. 
Table \ref{tab:EnsemblesSignificanceTable}  presents the statistical significance of the ranking difference between the examined ensembles and the actual best-classifier. 
Lastly, the statistical significance of the ranking difference between ensembles and ensemble-members are presented in Table \ref{tab:MembersSignificanceTable}.


\begin{table}[h]
	\centering
		\resizebox{1\linewidth}{!} {
		\begin{tabular}{|@{\hspace{0.5mm}}l@{\hspace{0.5mm}}|p{0.26\linewidth}@{}p{0.2\linewidth}@{}c@{\hspace{2mm}}p{0.18\linewidth}@{}c@{\hspace{2mm}}p{0.2\linewidth}@{}cc@{\hspace{1mm}}|}
		
\noalign{\hrule height 1pt}
														& &	Random \scriptsize{Classifier}	&	\multirow{2}{*}{ESBE} &	Majority \newline Voting	&	\multirow{2}{*}{Max}	&	Mean \newline Voting	&	\multirow{2}{*}{Product}	&	\multirow{2}{*}{TUPSO} \\
\noalign{\hrule height 1pt}
\multirow{34}{*}{\begin{sideways}UCI Repository\end{sideways}}
& Anneal 						& 0.648 (4) & 0.624 (5) & 0.505 (6) & 0.500 (7) & 0.871 (2) & 0.874 \myem{(1)} & 0.869 (3) \\ 
& Audiology 				& 0.765 (5) & 0.857 (2) & 0.825 (3) & 0.532 (7) & 0.737 (6) & 0.798 (4) & 0.888 \myem{(1)} \\ 
& Balance-scale 		& 0.712 (5) & 0.783 (2) & 0.719 (3) & 0.517 (7) & 0.705 (6) & 0.716 (4) & 0.934 \myem{(1)} \\ 
& Breast-cancer 		& 0.495 (5) & 0.506 \myem{(1)} & 0.499 (4) & 0.500 (3) & 0.469 (7) & 0.476 (6) & 0.504 (2) \\ 
& C.H. disease 			& 0.560 (4) & 0.490 (7) & 0.512 (5) & 0.500 (6) & 0.646 (3) & 0.685 \myem{(1)} & 0.654 (2) \\ 
& Credit-rating 		& 0.638 (5) & 0.691 (4) & 0.598 (6) & 0.502 (7) & 0.721 (3) & 0.770 (2) & 0.788 \myem{(1)} \\ 
& E-coli 						& 0.917 (4) & 0.934 (2) & 0.932 (3) & 0.884 (6) & 0.819 (7) & 0.913 (5) & 0.963 \myem{(1)} \\ 
& H. Statlog 				& 0.569 (4) & 0.500 (6) & 0.520 (5) & 0.500 (7) & 0.640 (3) & 0.705 \myem{(1)} & 0.692 (2) \\ 
& Hepatitis 				& 0.615 (5) & 0.669 (4) & 0.579 (6) & 0.500 (7) & 0.684 (3) & 0.777 \myem{(1)} & 0.745 (2) \\ 
& Horse-Colic 			& 0.551 (5) & 0.552 (4) & 0.529 (6) & 0.502 (7) & 0.620 (3) & 0.622 (2) & 0.635 \myem{(1)} \\ 
& H. Disease 				& 0.619 (4) & 0.543 (6) & 0.573 (5) & 0.505 (7) & 0.749 (3) & 0.754 (2) & 0.791 \myem{(1)} \\ 
& Hypothyroid		 		& 0.507 (5) & 0.607 \myem{(1)} & 0.490 (7) & 0.496 (6) & 0.590 (3) & 0.593 (2) & 0.568 (4) \\ 
& Ionosphere 				& 0.846 (4) & 0.884 (3) & 0.898 (2) & 0.756 (6) & 0.682 (7) & 0.787 (5) & 0.964 \myem{(1)} \\ 
& Iris 							& 0.948 (4) & 0.904 (5) & 0.965 (3) & 0.975 (2) & 0.850 (7) & 0.895 (6) & 0.995 \myem{(1)} \\ 
& Chess 						& 0.622 (5) & 0.666 (4) & 0.529 (6) & 0.506 (7) & 0.773 (3) & 0.819 \myem{(1)} & 0.812 (2) \\ 
& Letter 						& 0.762 (5) & 0.876 (3) & 0.830 (4) & 0.502 (7) & 0.752 (6) & 0.906 (2) & 0.958 \myem{(1)} \\ 
& MFeature 					& 0.815 (6) & 0.924 (2) & 0.830 (5) & 0.608 (7) & 0.871 (4) & 0.873 (3) & 0.972 \myem{(1)} \\ 
& Mushroom 					& 0.674 (4) & 0.718 (3) & 0.574 (6) & 0.507 (7) & 0.645 (5) & 0.808 (2) & 0.881 \myem{(1)} \\ 
& Opt-Digits 				& 0.842 (5) & 0.927 (2) & 0.871 (4) & 0.658 (7) & 0.768 (6) & 0.897 (3) & 0.979 \myem{(1)} \\ 
& Page-Blocks 			& 0.768 (5) & 0.858 (3) & 0.848 (4) & 0.528 (7) & 0.752 (6) & 0.863 (2) & 0.944 \myem{(1)} \\ 
& Pen Digits 				& 0.936 (4) & 0.976 (3) & 0.985 (2) & 0.898 (6) & 0.834 (7) & 0.906 (5) & 0.997 \myem{(1)} \\ 
& Diabetes 					& 0.527 (4) & 0.493 (7) & 0.507 (5) & 0.505 (6) & 0.544 (3) & 0.552 (2) & 0.555 \myem{(1)} \\ 
& P-Tumor 					& 0.964 (5) & 0.975 (4) & 0.983 (3) & 1.000 \myem{(1)} & 0.876 (7) & 0.905 (6) & 1.000 \myem{(1)} \\ 
& Segment 					& 0.615 (5) & 0.662 (4) & 0.568 (6) & 0.534 (7) & 0.685 (3) & 0.704 (2) & 0.717 \myem{(1)} \\ 
& Sonar 						& 0.522 (4) & 0.473 (7) & 0.508 (5) & 0.495 (6) & 0.553 (3) & 0.583 \myem{(1)} & 0.575 (2) \\ 
& Soybean 					& 0.589 (4) & 0.524 (5) & 0.509 (6) & 0.503 (7) & 0.798 (2) & 0.784 (3) & 0.801 \myem{(1)} \\ 
& SPAM-Base 				& 0.584 (4) & 0.676 \myem{(1)} & 0.598 (3) & 0.500 (5) & 0.500 (5) & 0.500 (5) & 0.630 (2) \\ 
& Splice 						& 0.738 (5) & 0.976 (2) & 0.624 (6) & 0.500 (7) & 0.769 (4) & 0.862 (3) & 0.989 \myem{(1)} \\ 
& Vehicle 					& 0.608 (4) & 0.595 (5) & 0.525 (6) & 0.497 (7) & 0.731 (3) & 0.793 \myem{(1)} & 0.787 (2) \\ 
& Vote 							& 0.746 (4) & 0.923 (2) & 0.744 (5) & 0.504 (7) & 0.690 (6) & 0.830 (3) & 0.937 \myem{(1)} \\ 
& Vowel 						& 0.671 (4) & 0.651 (5) & 0.494 (7) & 0.500 (6) & 0.797 (3) & 0.813 (2) & 0.849 \myem{(1)} \\ 
& Waveform 					& 0.769 (4) & 0.752 (6) & 0.758 (5) & 0.661 (7) & 0.812 (3) & 0.814 (2) & 0.865 \myem{(1)} \\ 
& WB-Cancer 				& 0.768 (5) & 0.967 (2) & 0.953 (3) & 0.500 (7) & 0.725 (6) & 0.826 (4) & 0.979 \myem{(1)} \\ 
& Zoo 							& 0.732 (5) 	& 0.879 (2) & 0.656 (6) & 0.5 (7) & 0.756 (4) & 0.828 (3) & 0.943 \myem{(1)} \\ 
\noalign{\hrule height 1pt}
\multirow{6}{*}{\begin{sideways}KDD CAP 99\end{sideways}}
& AUTH 							& 0.913 (3) & 0.784 (7) & 0.978 (2) & 0.869 (5) & 0.861 (6) & 0.907 (4) & 0.989 \myem{(1)} \\ 
& FTP 							& 0.684 (5) & 0.730 (4) & 0.549 (6) & 0.509 (7) & 0.801 (3) & 0.883 (2) & 0.903 \myem{(1)} \\ 
& FTP-DATA 					& 0.655 (5) & 0.781 (2) & 0.602 (6) & 0.506 (7) & 0.713 (4) & 0.794 \myem{(1)} & 0.775 (3) \\ 
& OTHER 						& 0.825 (5) & 0.967 (2) & 0.748 (6) & 0.600 (7) & 0.895 (4) & 0.897 (3) & 0.995 \myem{(1)} \\ 
& POP3 							& 0.874 (5) & 0.964 (3) & 0.983 (2) & 0.750 (7) & 0.788 (6) & 0.909 (4) & 0.983 \myem{(1)} \\ 
& SMTP 							& 0.813 (4) & 0.970 \myem{(1)} & 0.737 (7) & 0.749 (5) & 0.748 (6) & 0.869 (3) & 0.931 (2) \\ 
\hline
\multicolumn{2}{|l}{Average Rank}		& 4.5 (5) & 3.6 (3) & 4.8 (6) & 6.3 (7) & 4.5 (5) & 2.9 (2) & \myem{1.4 (1)} \\
\noalign{\hrule height 1pt}

		\end{tabular}
		}
	\caption{Ensembles classification result table. Inside the parenthesis is the AUC rank of the tested classifier.}
	\label{tab:EnsembleResultTable}
\end{table}

The results in Table \ref{tab:EnsembleResultTable}, show that the ensemble with the highest average rank is the meta-learning based, TUPSO. The next best ensemble, by a large margin, is the product-rule, which was the only fixed-rule that was ranked higher than the random-classifier scheme. This is an indication for the high independence among the ensemble participants, induced by the heterogeneous learning algorithms. Indeed, \cite{Tax01combiningone-class} showed that the product-rule is motivated by independence of the combined models. In that work, however, the authors applied feature-set partitioning to decrease the dependency among the combined models.
The heuristic ensemble, ESBE, was ranked, on average, higher than the random classifier, showing, along with TUPSO, the benefits of using classification performance estimation. This simple ensemble was also ranked higher than the majority-voting, max-rule, and the mean-voting.

\begin{table}[b]
	\centering
		\resizebox{1\linewidth}{!} {
		\begin{tabular}{@{}p{0.26\linewidth}@{\hspace{1mm}}l@{\hspace{1mm}}l@{\hspace{1mm}}l@{\hspace{1mm}}l@{\hspace{1mm}}l@{\hspace{1mm}}l@{\hspace{1mm}}l@{}}
		\hline
		& \multicolumn{1}{c}{Majority} 	& \multicolumn{1}{c}{Max}	& \multicolumn{1}{c}{Mean} 	 & \multicolumn{1}{c}{Product}  & \multicolumn{1}{c}{\multirow{2}{*}{ESBE}} & \multicolumn{1}{c}{Random}			& \multicolumn{1}{c@{}}{Best} \\
		& \multicolumn{1}{c}{Voting} 		& \multicolumn{1}{c}{Rule} & \multicolumn{1}{c}{Voting} & \multicolumn{1}{c}{Rule} &  & \multicolumn{1}{c}{Classifier} & \multicolumn{1}{c@{}}{
Classifier} \\	
		\hline

		TUPSO & \myem{+} ($\approx 0$) & \myem{+} ($\approx 0$) & \myem{+} ($\approx 0$) & \myem{+} ($<0.01$) & \myem{+} ($<0.01$) & \myem{+} ($\approx 0$) &	\textbf{ \: (0.243)} \\  
	
	\cline{1-8}
	
		Best Classifier	& \myem{+} ($\approx 0$) & \myem{+} ($\approx 0$) & \myem{+} ($\approx 0$) & \myem{+} ($<0.01$) &  \myem{+} ($\approx 0$) & \myem{+} ($\approx 0$) &  \vspace{1mm} \\ \cline{1-7}
		
		
		Rand Classifier	&  \:\: (0.786) & \myem{+} ($<0.01$) &  \:\: (0.964) & $\myem{-}$ ($<0.01$) & $ \:\:\:$ (0.099) &  &  \vspace{1mm} \\ \cline{1-6}
		
		ESBE				& \myem{+} ($0.057$) & \myem{+} ($\approx 0$) & \:\: (0.108) & \:\:\:\:\: (0.152) &  &  &  \\ 
		\hline
		\end{tabular}
		}
	\caption{The statistical significance of the difference in the AUC measure of the examined ensembles. The $p$-$value$ of the test statistic is inside the brackets.}
	\label{tab:EnsemblesSignificanceTable}
\end{table}
The statistical tests in Table \ref{tab:EnsemblesSignificanceTable} reveal four clusters of classifiers, each comprised of statistically comparable ensembles or classifiers. TUPSO and the best-base classifier populate the cluster that represents the top-tier classification performance. ESBE and product-rule ensembles comprise the cluster that represents the above-average classification performance. The majority-voting, mean voting and randomly selected classifier make the below-average performance cluster, and finally, the cluster that represents the lowest classification performance is comprised of the max-rule ensemble.


\begin{table}[h]
	\centering
		\resizebox{1\linewidth}{!} {
		\begin{tabular}{@{}p{0.23\linewidth}@{\hspace{0.8mm}}p{0.21\linewidth}@{}p{0.21\linewidth}@{\hspace{0.0mm}}l@{\hspace{0.5mm}}l@{\hspace{0.5mm}}l@{\hspace{0.5mm}}p{17mm}@{}}
		\hline
		& \multicolumn{6}{c}{Base Classifiers (ensemble members)} \\ \cline{2-7}
		&	 \multicolumn{1}{c}{ADIFA$_{GM}$}	&	\multicolumn{1}{c}{ADIFA$_{HM}$}	&	\multicolumn{1}{c}{GDE} &	\multicolumn{1}{c}{PGA} &	\multicolumn{1}{c}{OC-SVM$_1$} &	\multicolumn{1}{c}{OC-SVM$_2$} \\
		\hline
		TUPSO 			& \myem{+} ($<0.01$) & \myem{+} ($<0.01$) & \myem{+} (0.043) &  \myem{+} ($\approx 0$) & \myem{+} ($\approx 0$)  & \myem{+} ($\approx 0$) \\
		\hline
		ESBE 				&  \multicolumn{1}{c}{(0.681)} &  \multicolumn{1}{c}{(0.719)} &  \multicolumn{1}{c}{(0.148)} &   \multicolumn{1}{c}{(0.344)} & \myem{+} ($<0.01$)  & \myem{+} ($<0.01$) \\
		\hline
		Maj. Voting					&  \multicolumn{1}{c}{(0.203)} &  \multicolumn{1}{c}{(0.129)} & $\myem{-}$ ($0.011$) &  \multicolumn{1}{c}{(0.590)} & \myem{+} ($<0.01$)  & \myem{+} ($<0.01$) \\
		\hline
		Mean Vote					&  \multicolumn{1}{c}{(0.489)} &  \multicolumn{1}{c}{(0.141)} & $\myem{-}$ ($<0.01$) &   \multicolumn{1}{c}{(0.797)} & \myem{+} ($<0.01$)  & \myem{+} ($<0.01$) \\
		\hline
		Product			&  \multicolumn{1}{c}{(0.555)} &  \multicolumn{1}{c}{(0.918)} &  \multicolumn{1}{c}{(0.170)} &   \multicolumn{1}{c}{(0.284)} & \myem{+} ($<0.01$)  & \myem{+} ($<0.01$) \\
		\hline
		Best Clas. 	& \myem{+} ($\approx 0$)  & \myem{+} ($\approx 0$)  & \myem{+} ($<0.01$)   &  \myem{+} ($\approx 0$)  & \myem{+} ($\approx 0$) & \myem{+} ($\approx 0$)  \\
		\hline
		\end{tabular}
		}
	\caption{The statistical significance of the difference in the AUC measure between the examined ensembles and their ensemble-members.} 
	\label{tab:MembersSignificanceTable}
\end{table}

Table \ref{tab:MembersSignificanceTable} indicates that the average AUC of TUPSO was significantly higher than that of all the ensemble-members. In addition, the table shows that the base-classifier with the highest average AUC measure, i.e., GDE, achieved a significantly lower AUC than the actual best classifier. This fact further supports our claim regarding the inefficiency of choosing the best classifier by solely relying on its classification performance on previous classification tasks.

\begin{figure}[t]
	\centering
	\includegraphics[scale=0.315]{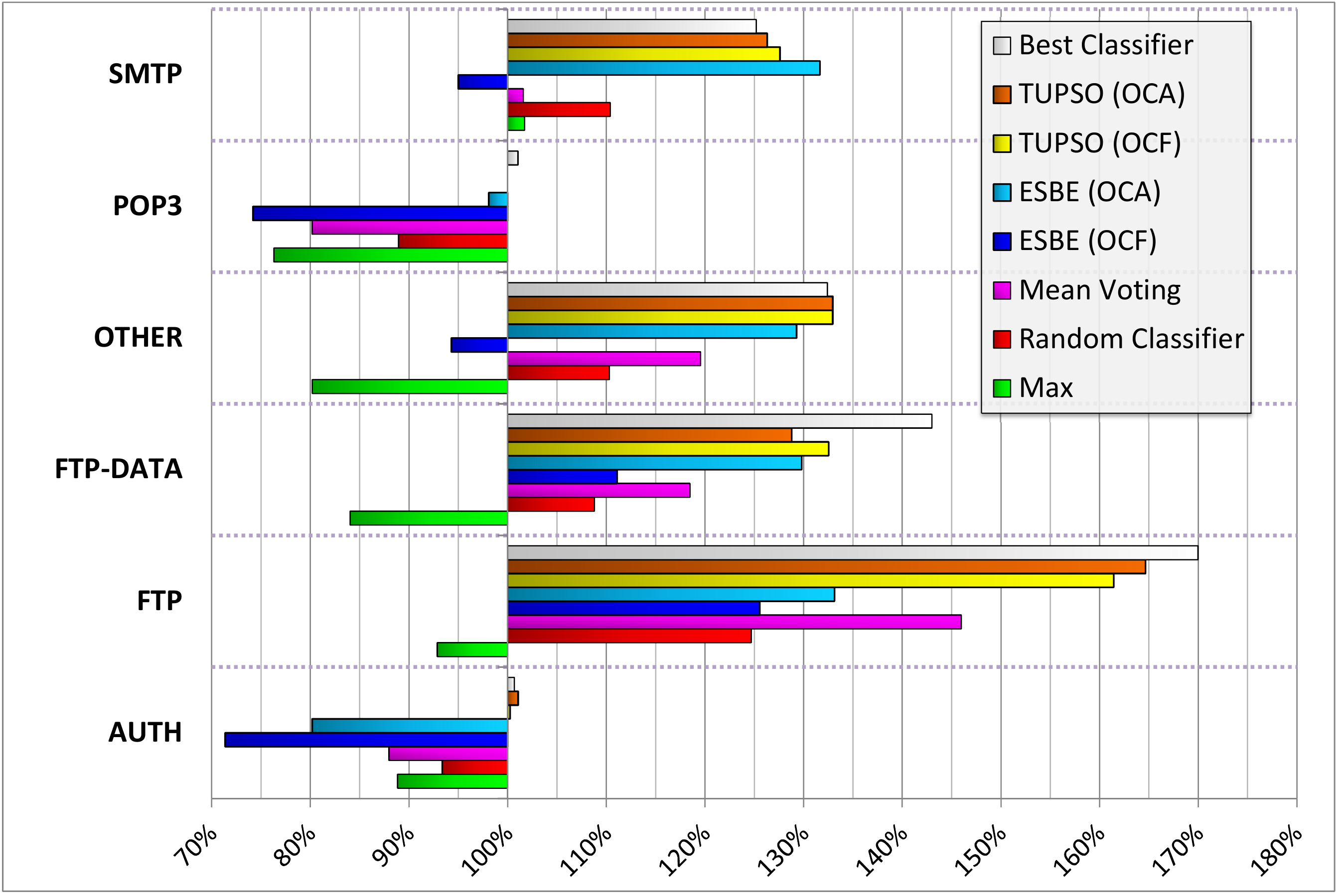}
	\caption{Comparing the ensembles` performance. The majority-voting is the reference ensemble.}
	\label{fig:comparingToVotingPOA}
\end{figure}

Figures \ref{fig:comparingToVotingPOA} presents the AUC performance of different ensembles relative to the AUC of the most popular one-class combining method, i.e., Majority voting,  on the KDD-CUP-99 datasets. The figure shows that relatively to TUPSO, the ESBE ensemble is affected much more by the type of performance estimator. This may indicate that the TUPSO's meta-features are only mildly influenced by the performance estimator.

\subsection{Classifying \emph{Like} the Best-Classifier?}

We continue further with our experiment to find out which of the ensembles can be used as an alternative for the best base-classifier. Assuming that the users require a one-class classifier for a single classification task (i.e., a single dataset), they will be more concerned by how well their classifier will perform on their classification task, rather than how it will perform on average (i.e., on many different datasets).
In our case, the best performer, .i.e, TUPSO, is on average as good as the best base-classifier and therefore is the natural choice of ensemble. However, it might be the case that on several datasets TUPSO significantly outperforms the best base-classifier, while on other datasets, it significantly falls behind the best base-classifier. If this is the case, one might want to use another ensemble that has a greater chance of classifying like the best base-classifier.

\begin{figure}[h]
	\centering
		\includegraphics[scale=0.31]{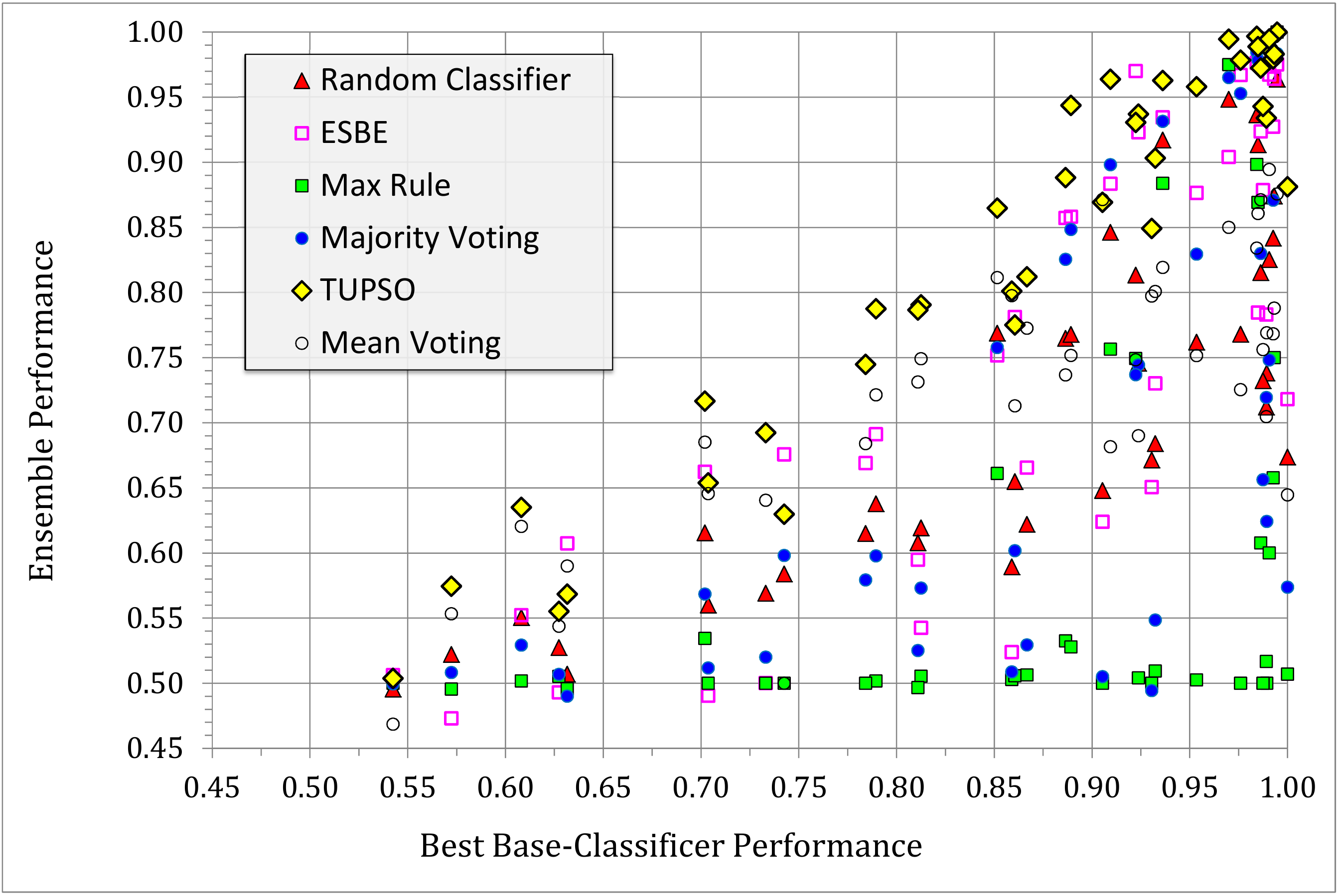}
	\caption{Classification performance: ensembles vs. actual best classifier.}
	\label{fig:Correlation1}
\vspace{-1mm}
\end{figure}

Figure \ref{fig:Correlation1} graphically shows the link between the AUC performance of the best base-classifier 
and that of the ensembles. 
Each dataset is represented by a single data point. To find out which of the ensembles` AUC performance is most tightly linked with the best base-classifier, we computed a correlation matrix using the Pearson Correlation routine. The results in Table \ref{tab:PearsonCorrelation} show that the TUPSO ensemble has the most correlated AUC performance with the best base-classifier.

\begin{table}[h!]
	\centering
		\resizebox{1\linewidth}{!} {
		\begin{tabular}{p{0.20\linewidth}p{0.17\linewidth}p{0.17\linewidth}p{0.17\linewidth}p{0.17\linewidth}p{0.17\linewidth}p{0.17\linewidth}p{0.17\linewidth}}
			\hline
			& Random Classifier & Majority Voting & Max & Mean \newline Voting & Product & ESBE & TUPSO \\
			\hline
			Pearson \newline Correlation & \multicolumn{1}{c}{\multirow{2}{*}{\cellcolor[rgb]{1.000,0.921,0.517} 0.815}}
																	 & \multicolumn{1}{c}{\multirow{2}{*}{\cellcolor[rgb]{0.988,0.709,0.474} 0.661}}
																	 & \multicolumn{1}{c}{\multirow{2}{*}{\cellcolor[rgb]{0.972,0.415,0.415} 0.439}} 
																	 & \multicolumn{1}{c}{\multirow{2}{*}{\cellcolor[rgb]{1.000,0.882,0.510} 0.788}} 
																	 & \multicolumn{1}{c}{\multirow{2}{*}{\cellcolor[rgb]{0.717,0.834,0.500} 0.883}} 
																	 & \multicolumn{1}{c}{\multirow{2}{*}{\cellcolor[rgb]{0.717,0.834,0.500} 0.812}} 
																	 & \multicolumn{1}{c}{\multirow{2}{*}{\cellcolor[rgb]{0.399,0.900,0.480} 0.962}}\\
		
			\hline
		\end{tabular}
		}
	\caption{Pearson Correlation.}
	\label{tab:PearsonCorrelation}
\end{table}
\subsection{One-Class Performance Estimation \newline \& Meta-Learning}
In this last experiment we examine the contribution of the performance evaluators, OCA and OCF, to the classification performance of the meta-learning based ensemble, TUPSO.
To do so, we experimented with TUPSO on two different Meta features types separately, i.e., the summation-based Meta features ($f_2$ and $f_3$) and the variance-based Meta features ($f_7$ and $f_8$). For each meta-features type, three TUPSO executions were made 
; The first, without any weighting, for the second and third executions, OCA and OCF were respectively applied in order to calculate the ensemble's members’ performance. Mind that in each experiment, only a single meta-feature was used. 
In this experiment we execute TUPSO on 40 UCI datasets and use the OC-SVM learning algorithm for training the meta-classifier. The experimental results are presented in Table \ref{tab:TUPSOAndHueristicMetrics}.

\begin{table}[h!]
	\centering
		\resizebox{1\linewidth}{!} {
		\begin{tabular}{|@{\hspace{1mm}}p{0.36\columnwidth} | c!{\color{gray}\vrule} c c | c!{\color{gray}\vrule} c c|}
			\hline
			\scriptsize \raggedleft Meta feature type:	& \multicolumn{3}{c|}{Sum of Predictions}	& \multicolumn{3}{c|}{Variance of Predictions} \\
			\hline
			\scriptsize \raggedleft  Meta Feature: & $f_2$	& \multicolumn{2}{c|}{$f_3$}	& $f_7$	& \multicolumn{2}{c|}{$f_8$} \\ \arrayrulecolor{gray} \cline{2-7}
			\arrayrulecolor{black}
			Dataset  \hfill \scriptsize Perf. Measure: & none	& OCA	& OCF	& none	& OCA	& OCF \\
			\hline
			Anneal & 0.5 (3) & 0.537 (2) & 0.61 \myem{(1)} & 0.52 (3) & 0.534 (2) & 0.538 \myem{(1)} \\
			Arrhythmia & 0.488 \myem{(1)} & 0.474 (3) & 0.474 (2) & 0.527 \myem{(1)} & 0.479 (3) & 0.484 (2) \\
			Audiology & 0.644 \myem{(1)} & 0.633 (2.5) & 0.633 (2.5) & 0.508 (3) & 0.57 \myem{(1)} & 0.539 (2) \\
			Balance-scale & 0.542 (3) & 0.73 \myem{(1)} & 0.724 (2) & 0.65 (3) & 0.797 (2) & 0.802 \myem{(1)} \\
			Breast-cancer & 0.509 (3) & 0.514 \myem{(1)} & 0.51 (2) & 0.514 \myem{(1)} & 0.503 (3) & 0.512 (2) \\
			Heart-Disease & 0.5 (3) & 0.535 \myem{(1)} & 0.512 (2) & 0.496 (3) & 0.506 (2) & 0.531 \myem{(1)} \\
			Credit Rating & 0.556 (3) & 0.613 (2) & 0.666 \myem{(1)} & 0.518 (3) & 0.527 (2) & 0.583 \myem{(1)} \\
			E-Coli & 0.882 (3) & 0.905 (2) & 0.905 \myem{(1)} & 0.466 (3) & 0.483 \myem{(1)} & 0.473 (2) \\
			Glass & 0.515 (2) & 0.515 (2) & 0.515 (2) & 0.512 \myem{(1)} & 0.508 (2.5) & 0.508 (2.5) \\
			Heart Statlog & 0.5 (3) & 0.531 (2) & 0.536 \myem{(1)} & 0.501 (3) & 0.528 (2) & 0.546 \myem{(1)} \\
			Hepatitis & 0.55 (3) & 0.563 (2) & 0.565 \myem{(1)} & 0.547 (3) & 0.575 (2) & 0.583 \myem{(1)} \\
			Horse Colic & 0.529 (3) & 0.532 (2) & 0.538 \myem{(1)} & 0.513 \myem{(1)} & 0.505 (3) & 0.509 (2) \\
			Heart-Disease-(H) & 0.52 (3) & 0.622 \myem{(1)} & 0.604 (2) & 0.56 (2) & 0.558 (3) & 0.595 \myem{(1)} \\
			Thyroid Disease & 0.497 \myem{(1)} & 0.485 (3) & 0.485 (2) & 0.509 (2) & 0.499 (3) & 0.51 \myem{(1)} \\
			Ionosphere & 0.885 (3) & 0.892 (2) & 0.893 \myem{(1)} & 0.522 \myem{(1)} & 0.51 (2) & 0.509 (3) \\
			Iris & 0.965 \myem{(1)} & 0.96 (2.5) & 0.96 (2.5) & 0.48 (3) & 0.5 \myem{(1)} & 0.495 (2) \\
			Chess  & 0.591 (1.5) & 0.516 (3) & 0.591 (1.5) & 0.595 \myem{(1)} & 0.574 (3) & 0.594 (2) \\
			Labor & 0.506 (3) & 0.556 (2) & 0.563 \myem{(1)} & 0.535 (3) & 0.59 \myem{(1)} & 0.582 (2) \\
			Letter & 0.5 (3) & 0.822 \myem{(1)} & 0.82 (2) & 0.516 (3) & 0.587 (2) & 0.75 \myem{(1)} \\
			Lymphography & 0.528 (3) & 0.538 \myem{(1)} & 0.532 (2) & 0.525 \myem{(1)} & 0.52 (2) & 0.515 (3) \\
			M-Features-Pixel & 0.724 (3) & 0.883 (2) & 0.884 \myem{(1)} & 0.59 \myem{(1)} & 0.566 (3) & 0.585 (2) \\
			Mushroom & 0.6 (2) & 0.561 (3) & 0.685 \myem{(1)} & 0.615 (3) & 0.672 (2) & 0.759 \myem{(1)} \\
			Opt Digits & 0.897 \myem{(1)} & 0.864 (3) & 0.867 (2) & 0.576 (3) & 0.58 \myem{(1)} & 0.578 (2) \\
			Page-Blocks & 0.599 (3) & 0.618 (1.5) & 0.618 (1.5) & 0.505 (3) & 0.547 \myem{(1)} & 0.547 (2) \\
			Pen Digits & 0.982 \myem{(1)} & 0.98 (2) & 0.979 (3) & 0.48 (3) & 0.492 (2) & 0.506 \myem{(1)} \\
			Pima Diabetes & 0.502 (3) & 0.531 \myem{(1)} & 0.511 (2) & 0.572 \myem{(1)} & 0.492 (3) & 0.518 (2) \\
			Primary Tumor & 0.517 (3) & 0.521 (1.5) & 0.521 (1.5) & 0.502 \myem{(1)} & 0.495 (3) & 0.5 (2) \\
			Segment & 0.923 (3) & 0.98 (2) & 0.981 \myem{(1)} & 0.482 (2) & 0.492 \myem{(1)} & 0.48 (3) \\
			Thyroid Disease II & 0.582 \myem{(1)} & 0.565 (3) & 0.574 (2) & 0.514 (3) & 0.524 (2) & 0.524 \myem{(1)} \\
			Sonar & 0.514 (3) & 0.517 (1.5) & 0.517 (1.5) & 0.478 (3) & 0.478 (2) & 0.484 \myem{(1)} \\
			Soybean & 0.498 (2) & 0.498 (3) & 0.508 \myem{(1)} & 0.516 (2) & 0.628 \myem{(1)} & 0.5 (3) \\
			Spambase & 0.594 \myem{(1)} & 0.584 (3) & 0.591 (2) & 0.486 (3) & 0.5 (1.5) & 0.5 (1.5) \\
			Splice & 0.69 (2) & 0.605 (3) & 0.967 \myem{(1)} & 0.929 (3) & 0.955 \myem{(1)} & 0.949 (2) \\
			Tic-Tac-Toe & 0.473 \myem{(1)} & 0.469 (3) & 0.472 (2) & 0.502 (3) & 0.548 (2) & 0.554 \myem{(1)} \\
			Vehicle & 0.565 (2) & 0.579 \myem{(1)} & 0.56 (3) & 0.636 (3) & 0.662 \myem{(1)} & 0.65 (2) \\
			Vote & 0.703 (3) & 0.74 (2) & 0.758 \myem{(1)} & 0.611 \myem{(1)} & 0.592 (3) & 0.605 (2) \\
			Vowel & 0.517 (3) & 0.625 (2) & 0.65 \myem{(1)} & 0.489 \myem{(1)} & 0.472 (3) & 0.486 (2) \\
			Waveform & 0.536 (3) & 0.809 (2) & 0.813 \myem{(1)} & 0.526 (3) & 0.551 (2) & 0.563 \myem{(1)} \\
			W-Breast-Cancer & 0.63 (3) & 0.897 (2) & 0.954 \myem{(1)} & 0.504 (2) & 0.498 (3) & 0.825 \myem{(1)} \\
			Zoo & 0.55 (3) & 0.563 (2) & 0.588 \myem{(1)} & 0.69 (3) & 0.721 (2) & 0.74 \myem{(1)} \\
\hline
Average AUC:    & 0.61 (2.39) & 0.65 (2.04) & \myem{0.67 (1.58)} & 0.54 (2.28) & 0.56 (2.05) & \myem{0.58 (1.68)} \\
 \hline
		\end{tabular}
		}
	\caption{The contribution of the ensemble-member`s performance estimation. Inside the parenthesis is the AUC rank of the tested method with respect to the tested feature.}
	\label{tab:TUPSOAndHueristicMetrics}
\end{table}

As can be seen in Table \ref{tab:TUPSOAndHueristicMetrics}, the AUC performance for two feature groups is improved by the weighting methods. It is also visible that the OCF performance measure produces better AUC values compared to the OCA by a noticeable margin. Next, we check whether the difference in the AUC is significant statistically. Table \ref{tab:TupsoSignificance} shows that when the ensemble-members' predictions are weighted by the OCF, the AUC results are significantly higher compared to when the predictions are not weighted.

\begin{table}[h!]
	\centering
	\resizebox{1\linewidth}{!} {
		\begin{tabular}{@{}l | c  c c  c c@{}}
		\hline
			& Weighted Feature: & \multicolumn{2}{c}{Sum-Predictions} & \multicolumn{2}{c@{}}{Var-Predictions} \\
			Non-weighted Feature: 						& Perf. Measure: & OCA & OCF & OCA & OCF  \\
		\hline
		
		Variance-Predictions &  						&  	$n/a$	& $n/a$ 		& (0.3143) & \myem{$+$} (0.007) \\ \cline{1-6}
		
		Sum-Predictions 			&  						& (0.1175) 	& \myem{$+$} (0.0003) &  		&  	 \\ \cline{1-4}
		
		\end{tabular}
		}
	\caption{The significance of the difference between weighted features and their non-weighted matched using the AUC metric. The \emph{p-value} of the test statistic is inside the brackets.}
	\label{tab:TupsoSignificance}
\end{table}

\section{Conclusions}
\label{sec:conclusions}
Thus far, the only combining scheme used for diverse learning algorithms in the context of one-class learning is the Fix-rule. Judging by publication quantity, schemes, such as majority voting, mean-voting, and max-rule are the most popular. However, in this study we hypothesized the existence of an innate limitation with such combining methods, as they do not take into account the properties of the ensemble-members they combine. Further along, we empirically demonstrated this limitation. The fixed-rule schemes indeed produced a lower classification accuracy when compared to the best base classifier.

In this paper we searched for an improved method for combining one-class classifiers. We implemented two one-class classification performance evaluators, OCA and OCF, which evaluated ensemble-members using the available positive labeled instances only. We then introduced ESBE, a simple ensemble that uses OCA or OCF to select a single classifier from the classifiers pool. Our experiment showed that while this method is inferior to the best ensemble-member, it still outperforms the Majority voting. 

Lastly, we introduced a meta-learning based ensemble, TU\-PSO, which learns a combining function upon aggregates of the ensemble-members' predictions. Thus, in contrast to the fix-rule scheme, TUPSO depends on the classification properties of the ensemble-members. Our experiments demonstrated the superiority of TUPSO over all other tested ensembles, both in terms of classification performance and in correlation to the best ensemble-member. Furthermore, TUPSO was found to be statistically indistinguishable from the best ensemble-member, thus, it completely removes the necessity of choosing the best ensemble-member.

\bibliographystyle{abbrv}
\bibliography{mybib}
\end{document}